\documentclass[letterpaper, 10 pt, conference]{ieeeconf}  

\usepackage{graphicx}
\usepackage{amsmath,amssymb,amsfonts}
\usepackage{multirow}
\usepackage{algorithm}
\usepackage{algorithmic}
\usepackage{psfrag}
\usepackage{subfigure}
\usepackage{color}
\usepackage{soul}
\usepackage{xcolor}
\usepackage[bookmarks=false]{hyperref}
\usepackage{lipsum}
\usepackage{mathrsfs}
\usepackage{pbox}
\usepackage{flushend}
\usepackage{mathtools}
\usepackage{tabularx}

\makeatletter
\def\hlinewd#1{%
  \noalign{\ifnum0=`}\fi\hrule \@height #1 \futurelet
   \reserved@a\@xhline}
\makeatother

\graphicspath{{Figures/}}

\IEEEoverridecommandlockouts                              

\title{\LARGE \bf
WatchPed: Pedestrian Crossing Intention Prediction\\ Using Embedded Sensors of Smartwatch
}

\author{Jibran Ali Abbasi$^{1*}$, Navid Mohammad Imran$^{2*}$, Lokesh Chandra Das$^{2}$ and Myounggyu Won$^{2}$
\thanks{$^{*}$Jibran Ali Abbasi and Navid Mohammad Imran have equally contributed to this work.}
\thanks{$^{2}$Jibran Ali Abbasi is with Afiniti, Virginia, United States
	 {\tt\small \{jibranabbasi.91\}@gmail.com}}
\thanks{$^{1}$Navid Mohammad Imran, Lokesh Chandra Das and Myounggyu Won are with the Department of Computer Science, University of Memphis, Memphis, TN, United States
        {\tt\small \{nimran, ldas, mwon\}@memphis.edu}}%
}

\begin{document}

\maketitle
\thispagestyle{empty}
\pagestyle{empty}

\begin{abstract}
The pedestrian crossing intention prediction problem is to estimate whether or not the target pedestrian will cross the street. State-of-the-art techniques heavily depend on visual data acquired through the front camera of the ego-vehicle to make a prediction of the pedestrian's crossing intention. Hence, the efficiency of current methodologies tends to decrease notably in situations where visual input is imprecise, for instance, when the distance between the pedestrian and ego-vehicle is considerable or the illumination levels are inadequate. To address the limitation, in this paper, we present the design, implementation, and evaluation of the first-of-its-kind pedestrian crossing intention prediction model based on integration of motion sensor data gathered through the smartwatch (or smartphone) of the pedestrian. We propose an innovative machine learning framework that effectively integrates motion sensor data with visual input to enhance the predictive accuracy significantly, particularly in scenarios where visual data may be unreliable. Moreover, we perform an extensive data collection process and introduce the first pedestrian intention prediction dataset that features synchronized motion sensor data. The dataset comprises 255 video clips that encompass diverse distances and lighting conditions. We trained our model using the widely-used JAAD and our own datasets and compare the performance with a state-of-the-art model. The results demonstrate that our model outperforms the current state-of-the-art method, particularly in cases where the distance between the pedestrian and the observer is considerable (more than 70 meters) and the lighting conditions are inadequate.
\end{abstract}


\section{Introduction}
\label{sec:introduction}

The pedestrian crossing intention prediction problem is to forecast whether or not the target pedestrian will cross the street over a short time horizon~\cite{zhao2021action}. Accurately predicting the pedestrian's crossing intention is of great importance for assuring traffic safety~\cite{quintero2017pedestrian,liang2019peeking}, especially for enhancing the safety of advanced driver assistance systems (ADAS) both for human-driven and autonomous vehicles~\cite{razali2021pedestrian}. Specifically, by providing early warnings, the driver or the autonomous vehicle itself can earn an ample time for responding to unexpected situations to prevent accidents~\cite{cadena2022pedestrian}. In fact, the development of effective methods for predicting pedestrian crossing intentions is a crucial prerequisite for achieving level-4 autonomy particularly for urban environments where autonomous vehicles frequently interact with pedestrians~\cite{yang2022predicting}.

There are mainly two research directions for pedestrian intention prediction: one involves predicting the pedestrian's crossing intention by estimating their trajectory~\cite{alahi2016social,lee2017desire,bartoli2018context,gupta2018social,sadeghian2019sophie,mangalam2020not}, and machine learning-based approaches to directly estimate the pedestrian’s crossing intention based on different types of input features~\cite{neogi2017context,saleh2019real,gujjar2019classifying}. Many studies have demonstrated that trajectory-based techniques can effectively determine the pedestrian's intention to cross the street. Nevertheless, as these methods necessitate the pedestrian's motion history, in some real-world situations where pedestrians are stationary on the sidewalk, waiting for an opportunity to cross the street, trajectory-based techniques may not be viable~\cite{razali2021pedestrian,zhao2021action}. 

In the early stages, many machine learning-based methods for pedestrian's crossing intention prediction relied on convolutional neural networks (CNNs)~\cite{rasouli2017they}. Subsequently, to enhance the precision of pedestrian's crossing intention prediction, spatial and temporal correlations between video frames were incorporated by adopting recurrent neural networks~\cite{kotseruba2020they,lorenzo2020rnn,rasouli2019pie}. Different types of input features were fused to achieve better performance such as pedestrian poses, bounding boxes, ego-vehicle information, and semantic segmentation maps~\cite{fang2018pedestrian, rasouli2020pedestrian, rasouli2020multi, rasouli2021bifold}. Recent studies integrated additional input features related to the road geometry and surrounding people~\cite{yang2022predicting}. A prevalent feature of most existing machine learning-based methodologies is that the prediction of a pedestrian's crossing intention heavily depends on visual data obtained from the front camera of the ego-vehicle. Thus, the accuracy of predictions can significantly decrease when visual input is imprecise, such as when the distance between the pedestrian and ego-vehicle is substantial, or when the lighting conditions are inadequate.

To overcome these inherent limitations, we introduce WatchPed, a pedestrian's crossing intention prediction model that is designed to make a more accurate and reliable prediction based on the combination of visual information and motion sensor data collected from the embedded sensors of the pedestrian's smartwatch. Specifically, we propose a novel neural network architecture that effectively integrates heterogeneous data types to achieve unprecedented performance in predicting pedestrian crossing intentions. The proposed architecture consists of (1) the vision branch that processes the local and global visual context information, (2) the non-vision branch that consolidates non-visual features including the ego-vehicle speed, the bounding box of the pedestrian, and the pose information of the pedestrian, and (3) the sensor branch that incorporates the pedestrian's motion sensor data in the decision making process. Multiple input features generated from these branches are integrated efficiently, with the objective of achieving dependable pedestrian intention prediction even in situations where visual data alone may not be adequate.

Moreover, we perform an extensive data collection process and introduce the first pedestrian intention prediction dataset that features synchronized heterogeneous data types, including motion sensor and visual data. Specifically, the dataset we propose contains a total of 255 video clips that encompass diverse distances and lighting conditions, which make it an ideal resource for developing pedestrian intention prediction solutions based on motion sensor data. We trained our model using both the widely-adopted JAAD dataset as well as our own dataset and compared the performance with a state-of-the-art method~\cite{yang2022predicting}. Our results indicate that our approach markedly enhances performance, particularly in scenarios where the distance between the pedestrian and ego-vehicle is considerable, and lighting conditions are inadequate.

The following is a summary of our contributions:

\begin{itemize}
	\item We introduce the first pedestrian intention prediction model that integrates motion sensor data collected from the embedded sensors of the pedestrian's smartwatch to overcome the inherent limitations of state-of-the-art approaches that rely heavily on visual information to make predictions.
	\item We conduct a large-scale data collection process and introduce the first pedestrian intention dataset that features synchronized visual and non-visual data, as well as motion sensor data.
	\item We conduct an extensive experiment to demonstrate that our  model outperforms a state-of-the-art approach especially when the distance between the pedestrian and ego-vehicle is very far, and/or  lighting conditions are not sufficient.
\end{itemize}

This paper is organized as follows. We review state-of-the-art pedestrian intention prediction methods in Section~\ref{sec:related_work}, followed by the details of our dataset in Section~\ref{sec:xxx_dataset}. We then present the design of the architecture of the proposed model in Section~\ref{sec:proposed_approach}. The results of our experimental study are presented in Section~\ref{sec:results}. We then conclude in Section~\ref{sec:conclusion}.

\section{Related Work}
\label{sec:related_work}

The pedestrian crossing intention prediction problem is to forecast whether or not a given pedestrian will cross the road~\cite{kotseruba2021benchmark}. Numerous machine learning architectures based on manifold data modalities have been proposed to solve the problem. Some early efforts capitalized on static traffic scenes and pedestrian walking actions represented by CNNs to predict the pedestrian's intention~\cite{rasouli2017they}. The performance limitations due to stationary input features were amended by utilizing a sequence of frames ~\cite{lorenzo2020rnn}. Additionally, multimodal methods were studied to incorporate different types of information such as trajectories of the pedestrian~\cite{rasouli2020pedestrian}, body parts of the pedestrian~\cite{cadena2019pedestrian}, and interactions between pedestrians and other agents~\cite{liu2020spatiotemporal}. 

Numerous variants of existing approaches have been studied. Generative models were used to predict scene representations more effectively~\cite{gujjar2019classifying,chaabane2020looking}. Singh and Suddamalla incorporated both the global (\emph{i.e.,} visual features of different objects surrounding the pedestrian) and local context information (\emph{i.e.,} visual features of the target pedestrian)~\cite{singh2021multi}. Chen \emph{et al.} considered the elements of the ego-scene to provision future context for motivation~\cite{chen2021visual}. Lorenzo \emph{et al.} proposed a self-attention alternative based on transformer architecture which effectively fused video and pedestrian kinematic data~\cite{lorenzo2021capformer}. Razali \emph{et al.} designed a neural network based on a 5-block ResNet-50 to generate a map for predicting the probability for each pixel that belongs to a pedestrian~\cite{razali2021pedestrian}. Yang \emph{et al.} adopted a 3D CNN to capture the key behavioral information before the pedestrian crosses the street instead of relying on the skeleton features which may not be accurate if the distance between the pedestrian and vehicle is far~\cite{yang2021crossing}. Zhao \emph{et al.} proposed a method that combines the complementary strengths of multi-modal data to improve the performance~\cite{zhao2021action}. The vision transformer was utilized to process sequential data more effectively. Cadena \emph{et al.} proposed a modified GCN-based model by adding the contextual information such as the vision information about the environment~\cite{cadena2022pedestrian}. Zhang \emph{et al.} utilized CCTV videos to effectively capture the key points of the pedestrian body~\cite{zhang2021pedestrian}. Yang \emph{et al.} proposed a hybrid model that is designed to fuse different features from various input sources including both the non-visual and visual information~\cite{yang2022predicting}. 

Despite decent performance for predicting the pedestrian's intention, existing works have vital limitations that need immediate attention before being adopted as a key safety component of autonomous vehicles. Significant dependence on visual information makes them hard to predict the pedestrian's intention when the distance between the pedestrian and ego-vehicle is far. The performance further degrades when the lighting conditions are not enough. Yet, even a slim glitch is not permissible as far as safety is concerned. Therefore, our approach is designed to achieve the goal of maximizing pedestrian safety in such circumstances where the visual information may not be accurate by incorporating with the motion sensor data of the pedestrian collected with the pedestrian's smartwatch in performing pedestrian intention prediction. Moreover, a novel pedestrian crossing intention dataset which contains time-synchronized pedestrian motion sensor data has been created and publicized. 

\section{Proposed Dataset}
\label{sec:xxx_dataset}

\subsection{State-of-the-Art Pedestrian Intention Dataset}
\label{sec:pedestrian_dataset}

One of the first publicly available pedestrian crossing intention dataset is the Joint Attention in Autonomous Driving (JAAD) dataset~\cite{rasouli2018joint}. It contains 346 short video snippets recorded with a dashboard camera under different weather conditions. The JAAD dataset was utilized to evaluate numerous solutions for pedestrian crossing intention prediction. Another widely used dataset is Trajectory Inference using Targeted Action priors Network (TITAN) ~\cite{malla2020titan}. Compared to JAAD, TITAN is a large-scale dataset with fine-grained labeling. It consists of 10 hours of video (700 short clips) containing 8,592 unique pedestrians recorded in Tokyo, Japan. Stanford-TRI Intent Prediction (STIP) is another well recognized pedestrian intention dataset~\cite{liu2020spatiotemporal}. A distinctive feature of the STIP dataset is that the video was captured using three cameras that cover the left, right, and front of the ego-vehicle. Integration of multiple camera angles of the STIP data set improves the effectiveness of pedestrian intention prediction. The authors of the JAAD dataset have released a new version named Pedestrian Intention Estimation (PIE)~\cite{rasouli2019pie}. The PIE dataset was recorded in Toronto, Canada. Compared to its predecessor, JAAD, the PIE dataset features more comprehensive labeling and incorporates additional information on ego-vehicle motion.

\subsection{Our Dataset}

We conducted video recording on the streets of Memphis, Tennessee to create our dataset. We considered typical scenarios involving pedestrian crossings, in which a pedestrian wearing a smartwatch is present, and a vehicle is approaching the pedestrian. We utilized the dashboard camera of the vehicle to capture visual information of the pedestrian in the scene. A total of 255 HD video snippets were recorded consisting of a total of 120,245 annotated frames of pedestrians. Each video clip, captured at 30 frames per second, has a duration ranging from 10 to 30 seconds. The vehicle speed was recorded using the iPhone Timestamp Camera App once every 30 frames (\emph{i.e.,} every second). The Sensor Box App of a smart watch was used to record  50 accelerometer and gyroscope readings along with the corresponding timestamps per second. The full dataset is available at (\url{https://tinyurl.com/pedestrian2023}).

Our dataset features several unique aspects when compared to state-of-the-art datasets. Firstly, we strategically recorded video clips in various scenarios with different distances between the pedestrian and ego-vehicle. We also ensured that the video snippets were recorded in a variety of lighting conditions (\emph{i.e.,} sunny, night, cloudy, and rainy conditions). Secondly, our dataset comprises synchronized heterogeneous input features, including visual, non-visual, and motion sensor data. To the best of our knowledge, our dataset is the first of its kind that includes synchronized motion sensor data collected from the smartwatches of pedestrians. Thirdly, all motion sensor data transmitted from a mobile device are anonymous since the vehicle does not require knowledge of pedestrian identities to predict their crossing intentions. Thus, our solution is free from privacy concerns.

\subsection{Data Processing and Labeling}
\label{sec:data_processing}

All data we gathered were archived in the style of the JAAD dataset~\cite{rasouli2018joint}. More specifically, each video frame is annotated with the bounding box and pose information of the pedestrian. Additionally, behavioral tags including actions such as walking, standing, crossing, looking, \emph{etc.} are provided as annotations for each pedestrian per frame. The attributes of each pedestrian including age, gender, direction of motion, clothing, crossing location, \emph{etc.} are also provisioned for each frame. Each video is labeled with information on weather conditions and time of day, and annotations are also provided for visible objects in the scene, such as crosswalks, stop signs, traffic lights, \emph{etc.} Our dataset includes accelerometer and gyroscope data collected from the smartwatch of the pedestrian as the motion sensor data. The motion sensor data are time-synchronized with the corresponding video frames. More precisely, GPS-based timestamps were used to implement time synchronization at a precision of 0.001 second.

\begin{figure}[t]
	\centering
	\includegraphics[width=.99\columnwidth]{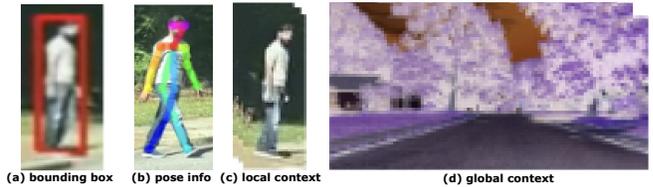}
	\caption {Examples of different types of input features provided in our dataset.}
	\label{fig:dataset_fig}
\end{figure}


To generate the bounding boxes of each pedestrian (\emph{e.g.,} see Fig.~\ref{fig:dataset_fig}(a)), the YOLOv4-1280 model (with DeepSort object tracking)~\cite{wojke2018deep} was used. This model is quite accurate in capturing bounding boxes when the distance between the pedestrian and ego-vehicle is close enough. However, it was observed that the model frequently failed to generate bounding boxes as the distance increased. 

To calibrate the YOLOv4-1240 model and successfully generate a bounding box for each and every frame, we used a linear interpolation method. More specifically, the Label Studio was used to manually produce bounding boxes that were not captured by the YOLOv4-1240 model. A sufficient number of bounding boxes were produced intermittently between frames. The gaps (\emph{i.e.,} frames with no bounding boxes between manually captured bounding boxes) were filled in through the linear interpolation method.


As we mentioned, each video frame is annotated with the pedestrian pose information (see Fig.~\ref{fig:dataset_fig}(b)). The OpenPose model~\cite{cao2017realtime} was used to capture the pose information. The model reads each frame and generates 18 key points for each pedestrian it detects. We observed that the model successfully captured pedestrian pose information when the pedestrian was in close proximity to the ego-vehicle. However, as the distance between the pedestrian and ego-vehicle increased, we encountered instances where non-human objects were incorrectly detected as humans and key points were generated for them. To fix this problem, we utilized the bounding box that was created in the previous step. More specifically, we purged the noisy pose information by incorporating only the key points that were within the pedestrian bounding box. Despite our efforts to improve the precision of the OpenPose model, the model failed to generate key points when the distance between the pedestrian and ego-vehicle was very far. For such frames, we did not provide annotations for pedestrian pose information.

Our dataset also provides both the local and global context by closely following the idea presented in~\cite{yang2022predicting}. The local context represents the visual features of the pedestrian. It comprises of a sequence of (224 $\times$ 224 pixels) RGB images of the surroundings of the pedestrian (see Fig.~\ref{fig:dataset_fig}(c)). The global context conveys the visual features of the objects in the video that can be used as a clue to infer the pedestrian-crossing activity. Pixel-level semantic masks were used to extract the global context from the video by classifying, localizing, and labeling those objects (see Fig.~\ref{fig:dataset_fig}(d)). DeepLabV3~\cite{chen2017rethinking} pre-trained with the Cityscapes Dataset~\cite{cordts2016cityscapes} was used to extract the semantic masks for the relevant objects. The extracted semantic segmentations were represented as (224 $\times$ 224 pixels) RGB images.

\section{Proposed Approach}
\label{sec:proposed_approach}

This section presents an overview of the proposed model, followed by the details of the model architecture. 

\subsection{Overview}
\label{sec:overview}

\begin{figure}[t]
	\centering
	\includegraphics[width=.9\columnwidth]{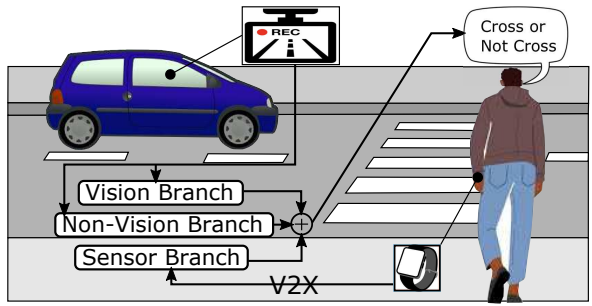}
	\caption {An operational overview of WatchPed.}
	\label{fig:overview}
\end{figure}

Fig.~\ref{fig:overview} displays an operational overview of WatchPed. An ego-vehicle makes a prediction of the pedestrian's crossing intention using a machine learning model trained with three different types of input feature vectors extracted from visual, non-visual, and motion sensor data, respectively. The visual features include the global and local environmental context. The non-visual features are the ego-vehicle speed, trajectory of the pedestrian (which is obtained based on the bounding box sequence of the pedestrian), and the pose key points of the pedestrian. The motion sensor features contain a sequence of accelerometer and gyroscope sensor data. As illustrated in the figure, visual and non-visual data are captured using the dashboard camera, and motion sensor data collected from the pedestrian's smartwatch is transmitted to the ego-vehicle via V2X (vehicle-to-everything) communication. The 5G V2P (vehicle to pedestrian) communication technology supports extremely low-latency and high-bandwidth wireless communication between vehicles and pedestrians~\cite{3gpp}. According to the recent standardization effort for 5G V2X~\cite{3gpp}, the peak data rate for 5G V2X is 20Gbps which is sufficient for supporting simultaneous communication with a large number of pedestrians to receive the sensor data from the pedestrians. 

Our goal is to design a neural network architecture for the ego-vehicle to estimate the probability distribution $\mathcal{P}(A_i^{t+f}| L_i, P_i, C_i, S_i, G, V) \in [0,1]$ that the target pedestrian $i$'s action $A_i^{t+f} \in \{0,1\}$ is to cross the street, where $t$ refers to the time of the latest frame, and $f$ is the number of frames representing the future time horizon for pedestrian intention prediction. Here, $L_i = \{l_i^{t-m}, l_i^{t-m+1}, ... l_i^t\}$ is the pedestrian $i$'s trajectory which is a sequence of the pedestrian $i$'s bounding box coordinates. More precisely, $l_i^{t-m} = \{x_{it}^{t-m}, y_{it}^{t-m}, x_{ib}^{t-m}, y_{ib}^{t-m}\}$, where $x_{it}^{t-m}, y_{it}^{t-m}$ are the x and y coordinates for the top-left point, and $x_{ib}^{t-m}, y_{ib}^{t-m}$ are the ones for the bottom-right point. $P_i = \{p_i^{t-m}, p_i^{t-m+1}, ... p_i^t\}$ refers to the pedestrian $i$'s pose information. Each element of $P_i$ represents a set of 18 2D coordinates for pose joints, \emph{i.e.,} $p_i^{t-m} = \{x_{i1}^{t-m}, y_{i1}^{t-m}, x_{i2}^{t-m}, y_{i2}^{t-m}, ..., x_{i18}^{t-m}, y_{i18}^{t-m}\}$. $G = \{g^{t-m}, g^{t-m+1}, ... g^t\}$ is the sequence of the global context, and $C_i = \{c_i^{t-m}, c_i^{t-m+1}, ... c_i^t\}$ is the sequence of the local context for pedestrian $i$.  $S_i = \{s_i^{t-m}, s_i^{t-m+1}, ... s_i^t\}$ is the sequence of the motion sensor data for pedestrian $i$. $V = \{v^{t-m}, v^{t-m+1}, ... v^t\}$ is the speed of the ego-vehicle. 

\subsection{Model Architecture}
\label{subsec:architecture}

\begin{figure*}[h]
	\centering
	\includegraphics[width=.9\textwidth]{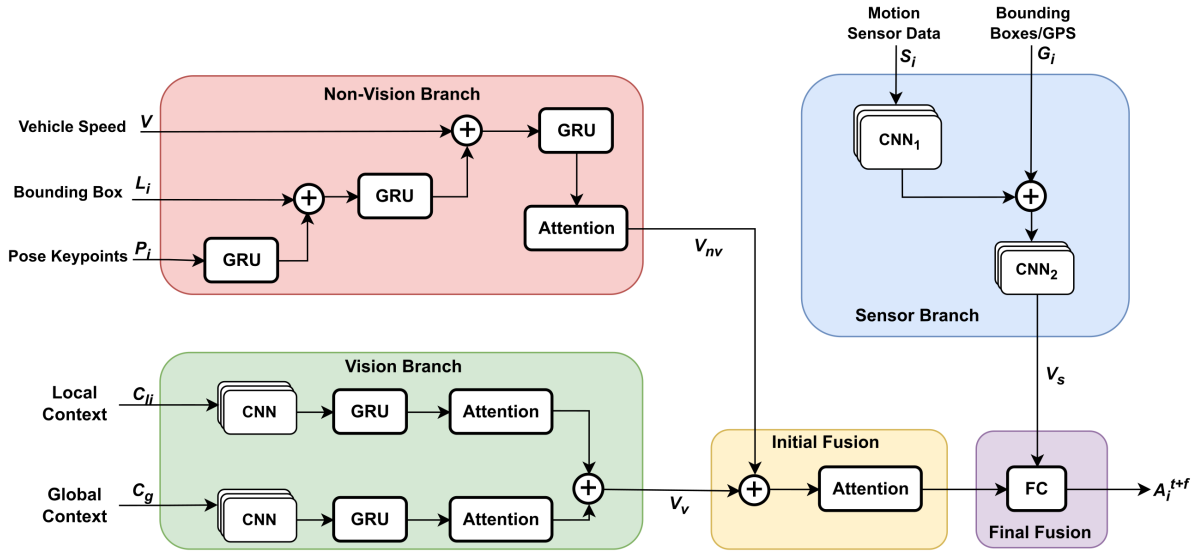}
	\caption {An architecture of the proposed pedestrian intention prediction model. An initial fusion is performed on the feature vectors produced by the non-vision and vision branches, and then the final fusion is performed by integrating with the output of the sensor branch.}
	\label{fig:architecture}
\end{figure*}

Fig.~\ref{fig:architecture} displays the neural network architecture for WatchPed. It consists of four main branches, \emph{i.e.,} the vision, non-vision, initial fusion, and motion sensor branches. The vision branch is composed of neural networks designed to process visual information to produce a vision feature vector $\mathbf{V_v}$. The non-vision branch consists of neural networks that generate a non-vision feature vector $\mathbf{V_{nv}}$ based on non-visual information as input. These two feature vectors are concatenated and provided as input to the attention block of the initial fusion branch. In parallel, the sensor branch processes the time-synchronized motion sensor data and pedestrian motion direction information to produce the motion sensor feature vector $\mathbf{V_{s}}$. The output of the sensor branch is subsequently fused with the output of the initial fusion branch to make a final prediction about the pedestrian $i$'s action $A_i^{t+f}$. 

We now present the detailed operation of each branch. The goal of the non-vision branch is to produce the non-vision feature vector $\mathbf{V_{nv}}$ using the temporal changes of the trajectory of the pedestrian, the key points of the pedestrian pose, and the ego-vehicle speed as input. More specifically, the pose information is provided to an RNN-based encoder, for which a gated recurrent unit (GRU)~\cite{cho2014properties} is adopted since a GRU is known to be more computationally efficient than LSTM. The GRU module consists of 256 hidden units. It outputs a feature tensor of size [16,256], which is then concatenated with the bounding box information (\emph{i.e.,} the trajectory information of the pedestrian). The concatenated vector is fed into another GRU module. The output is then concatenated with the vehicle speed information and is provided as input to the last GRU module. The output of the last GRU module is provisioned to the attention block, obtaining the non-vision vector $\mathbf{V_{nv}}$.

Similar to the non-vision branch, the vision branch produces the vision feature vector $\mathbf{V_{v}}$ using the local and global context as input. The local context represents the pedestrian's appearance around the bounding box. The global context captures ``significant'' objects in the semantically segmented scene. More specifically, the local context is processed by extracting spatial features using the CNN module and subsequently capturing the temporal features using the GRU module. For the CNN module, the VGG19 model~\cite{simonyan2014very} trained with the ImageNet dataset~\cite{deng2009imagenet} is adopted. An input to the CNN module is represented as a 4D array [number of observed frames, rows, cols, channels]. Following Yang \emph{et al.}'s work~\cite{yang2022predicting}, the dimension of the array was set to [16,244,244,3]. More precisely, the fourth maxpooling layer of the CNN module extracts the feature map of each local context image in the dimension of [512,14,14]. The pooling layer with a 14$\times$14 kernel then performs averaging of the feature map. And then, it is flattened and concatenated, producing the final feature tensor of size [16,512]. The spatial and temporal features for the global context are extracted in a similar manner based on the CNN and GRU modules. These features both for the local and global context are fed into the attention modules, respectively, and then they are concatenated to produce the final vision feature vector $\mathbf{V_{v}}$. The vision feature vector $\mathbf{V_{v}}$ is then fused with the non-vision feature vector $\mathbf{V_{nv}}$ in the initial fusion branch.

\renewcommand{\tabcolsep}{16.65pt}
\begin{table*}[t]
	\centering
	\caption{An effect of varying distances between pedestrian and ego-vehicle on performance of WatchPed and state-of-the-art model.}
	\begin{tabularx}{\textwidth}{ |c||c|c|c|c|c|c|}
		\hline
		\multirow{1}{*}{Distance to } & Models & Accuracy & AUC & F1-Score & Precision & Recall \\
		Pedestrian &  &  &  &  &  &  \\ \hline\hline
		\multirow{1}{*}{Close ($\sim$20m)} & Our Model & 0.8867 & 0.8522 & 0.9113 & 0.8664 & 0.9611 \\
		& State-of-the-Art~\cite{yang2022predicting} & 0.8562 & 0.8287 & 0.8812 & 0.8592 & 0.9045 \\
		\hline
		\multirow{1}{*}{Medium (20m$\sim$70m)} & Our Model & 0.7932 & 0.7856 & 0.7825 & 0.7866 & 0.7784 \\
		& State-of-the-Art~\cite{yang2022predicting} & 0.5416 & 0.5394 & 0.3927 & 0.5671 & 0.3004 \\
		\hline
		\multirow{1}{*}{\textbf{Far} (70m$\sim$180m)} & Our Model & \textbf{0.5924} & \textbf{0.6837} & \textbf{0.5549} & \textbf{1} & \textbf{0.384} \\
		& State-of-the-Art~\cite{yang2022predicting} & 0.3158 & 0.5 & 0 & 0 & 0 \\
		\hline
		\multirow{1}{*}{Average} & Our Model & 0.7574 & 0.7738 & 0.7495 & 0.8843 & 0.7078 \\
		& State-of-the-Art~\cite{yang2022predicting} & 0.5712 & 0.6227 & 0.4246 & 0.4754 & 0.4016 \\
		\hline
	\end{tabularx}
	\label{table:distance}
\end{table*}

The sensor branch processes the motion sensor data to produce the sensor feature vector $\mathbf{V_{s}}$. This branch is a crucial component of the proposed model that supports the limited decision-making capability of existing vision-based pedestrian intention prediction models. It allows maintaining high prediction accuracy in adverse situations where the pedestrian is located very far from the ego-vehicle, and the lighting conditions are poor, thereby enhancing the overall robustness of the model. More specifically, the motion sensor data in the form of a sequence of time-synchronized accelerometer and gyroscope sensor data are provided as input to a CNN module. This CNN module is designed to primarily detect whether the pedestrian is walking or jogging, which are precursors to the situation where the pedestrian is likely to cross the road. In essence, the CNN module is capable of promptly detecting the initial signal of a pedestrian intending to cross the street. However, relying solely on the walking or jogging motion may not be sufficient without taking into account the direction of the pedestrian's movement. Therefore, to improve the accuracy of pedestrian's crossing intention prediction, we integrate the output of the CNN module with input features that help infer the pedestrian's direction of movement, such as GPS coordinates and bounding boxes (Note here that, as we explained, these bounding boxes are pre-processed with our algorithm to enhance the detection rate even under very long distances to the pedestrian). And then, the combined input features are provided to another CNN module designed to capture the spatial feature related to the pedestrian moving direction. The output of the CNN module is then fused with the output of the initial fusion branch to generate the final expected action of the pedestrian. 

\section{Experimental Results}
\label{sec:results}

We evaluate the performance of WatchPed compared with a state-of-the-art method~\cite{yang2022predicting}. An experimental setting is presented in Section~\ref{sec:setup}, followed by an analysis of the experimental results that measures the performance with varying distances between the pedestrian and ego-vehicle in Section~\ref{subsec:impact_of_distance} and different levels of lighting conditions (Section~\ref{subsec:impact_of_lighting_conditions}).

\subsection{Experimental Setup}
\label{sec:setup}

\renewcommand{\tabcolsep}{7pt}
\begin{table}[t]
	\caption{Hyperarameters for the CNN modules of the sensor branch.}
	\begin{tabular}{ |c|p{2.5cm}|p{2.5cm}|}
		\hline
		& CNN$_1$ & CNN$_2$ \\ \hline
		Language & Python 3.10.4 & Python 3.10.4\\
		Model Type & TensorFlow (2.9.1) Sequential & TensorFlow (2.9.1) Sequential \\
		Frame Size & 100 & 60\\
		Hop Size & 50 & 10\\
		Epochs: & 15 & 100\\
		Test Size & 20\% & 20\%\\
		Optimizer & Adam &  Adam\\
		Learning Rate & 0.001 & 0.005\\
		Loss & sparse categorical crossentropy & sparse categorical crossentropy\\
		\hline
	\end{tabular}
	\label{table:sensor_branch_parameters}
\end{table}

We utilized a workstation running on Windows 10 OS equipped with an Intel Xeon Gold 5222 Processor, NVIDIA® RTX™ A4000, and 48GB RAM for training and testing both the proposed model and the state-of-the-art model~\cite{yang2022predicting}. The vision and non-vision branches were implemented by closely following~\cite{yang2022predicting}. Specifically, a dropout rate of 0.5 was used in the attention module. The L2 regularization of 0.001 was applied for the FC layer, and the Adam optimizer and the binary cross-entropy loss were used. Additionally, the learning rate of 5$\times10^{-7}$, batch size of 2, and 40 epochs were used. Table~\ref{table:sensor_branch_parameters} summarizes the hyperparameter values selected through an extensive trial and error process. The hyperparameter values were used to train the two CNN models of the sensor branch denoted by CNN$_1$ and CNN$_2$.

The state-of-the-art pedestrian intention prediction model~\cite{yang2022predicting} was trained using both the JAAD dataset and our dataset. Since the state-of-the-art model does not use motion sensor data to make predictions about pedestrian crossing intentions, the motion sensor data in our dataset were not included in training the state-of-the-art model. We also trained our model using both the JAAD dataset and our dataset. It is worth mentioning that our model is capable of making predictions even without the use of motion sensor data, which enables us to train it effectively using only the JADD dataset. To ensure a fair comparison, we evaluated both the state-of-the-art and our models using a common test dataset comprising 20\% randomly selected video clips from our dataset, which includes a range of distances and lighting conditions. The metrics we evaluated in this experiment include accuracy, area under the curve (AUC), F1-score, precision, and recall. We chose these metrics to ensure consistency with existing studies, as many previous pedestrian prediction solutions have used them to report their results~\cite{yang2022predicting}. 

\subsection{Effect of Distance}
\label{subsec:impact_of_distance}

\begin{figure*}[h]
	\centering
	\includegraphics[width=.99\textwidth]{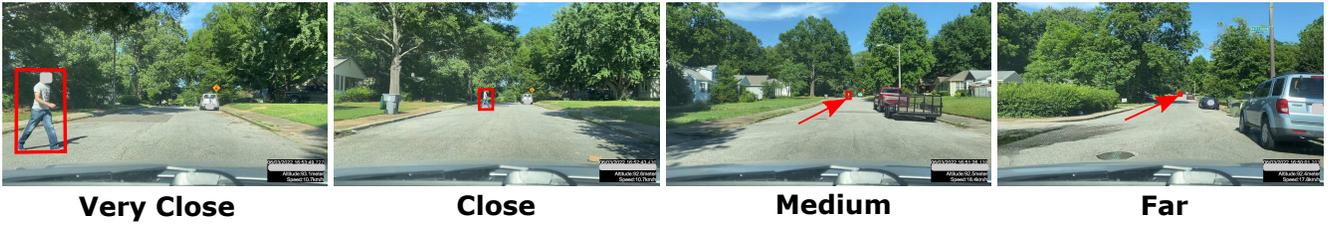}
	\caption {Examples of crossing pedestrians at different distances.}
	\label{fig:distance_example}
\end{figure*}

We evaluate the performance of our model in comparison with the state-of-the-art model by varying the distance between the pedestrian and ego-vehicle. To better understand the performance of our models at different distances, we organized the test dataset into three categories based on the distance to the pedestrian: close (approximately 20m), medium (20m-70m), and far (70m-180m). Fig.~\ref{fig:distance_example} displays snapshots of the pedestrian at different distances. The results are presented in Table~\ref{table:distance}. As the distance increased from close to far, we observed a degradation in both the accuracy and AUC of the state-of-the-art model by 63.1\% and 39.7\%, respectively. This suggests that the model's weakness lies in its inability to reliably interpret visual information at far distances. When the distance was far, we observed that the F1-score, precision, and recall for the state-of-the-art model were all 0, indicating that no positive observations were correctly predicted. This further highlights the model's weakness in accurately interpreting visual information at long distances.

\renewcommand{\tabcolsep}{17pt}
\begin{table*}[t]
	\centering
	\caption{An effect of varying lighting conditions on the performance of WatchPed and state-of-the-art model}
	\begin{tabularx}{\textwidth}{ |c||c|c|c|c|c|c|}
		\hline
		\multirow{1}{*}{Lighting Conditions} & Models & Accuracy & AUC & F1-Score & Precision & Recall \\
		&  &  &  &  &  &  \\ \hline\hline
		\multirow{1}{*}{Sunny} & Our Model & 0.6158 & 0.6155 & 0.6549 & 0.6282 & 0.6841 \\
		& State-of-the-Art~\cite{yang2022predicting} & 0.5032 & 0.5126 & 0.4875 & 0.5249 & 0.4552 \\
		\hline
		\multirow{1}{*}{Cloudy} & Our Model & 0.7647 & 0.7853 & 0.8007 & 0.9011 & 0.7205 \\
		& State-of-the-Art~\cite{yang2022predicting} & 0.5910 & 0.6389 & 0.5624 & 0.8356 & 0.4239 \\
		\hline
		\multirow{1}{*}{Rainy} & Our Model & 0.8645 & 0.8744 & 0.903 & 0.9564 & 0.8568 \\
		& State-of-the-Art~\cite{yang2022predicting} & 0.6876 & 0.7733 & 0.7709 & 0.9497 & 0.6488 \\
		\hline
		\multirow{1}{*}{\textbf{Night}} & Our Model & \textbf{0.7893} & \textbf{0.8265} & \textbf{0.8089} & \textbf{1} & \textbf{0.6792} \\
		& State-of-the-Art~\cite{yang2022predicting} & 0.6194 & 0.6942 & 0.5347 & 1 & 0.365 \\
		\hline
	\end{tabularx}
	\label{table:lighting}
\end{table*}


In contrast to the state-of-the-art model, our results demonstrate that our model is significantly more robust to changes in distance between the pedestrian and the ego-vehicle. While we observed a degradation in the performance of our model in terms of accuracy, AUC, F1-score, and recall as the distance increased, it still outperformed the state-of-the-art model across all distances. For far distances, our model showed a significant improvement over the state-of-the-art model, with accuracy and AUC increasing by up to 87.6\% and 36.7\%, respectively; not to mention, the F1-score, precision, and recall of our model are significantly higher when the distance is far since the values of the metrics for the state-of-the-art model were all 0. On average, our proposed model improved the accuracy, AUC, F1-score, precision, and recall by 32.6\%, 24.3\%, 76.5\%, 86.01\%, 76.2\%, respectively, compared with the state-of-the-art model. The result highlights the superior performance of our model in challenging scenarios where visual information is unreliable.

Interestingly, our model also showed slightly better performance than the state-of-the-art model even in scenarios where the distance was very close. This suggests that our model is not only more robust to changes in distance but also more effective at interpreting visual cues in close proximity. Specifically, our model improved the accuracy, AUC, F1-score, precision, and recall in ``close'' scenarios by 3.6\%, 2.8\%, 3.4\%, 0.8\%, and 6.2\%, respectively. A possible explanation is that in some cases the pedestrian pose information was not accurate even if the distance was very close, and the motion sensor data were used to rectify the inaccurate visual information, thereby improving the performance. 

\subsection{Effect of Lighting Conditions}
\label{subsec:impact_of_lighting_conditions}

We analyze the performance of our model compared with the state-of-the-art model by varying the lighting conditions. The test dataset was organized, based on the lighting conditions, into [sunny, cloudy, rainy, night]. The results are summarized in Table~\ref{table:lighting}. Interestingly, we made a counter-intuitive observation. More specifically, we hypothesized that the accuracy of the state-of-the-art model would be lower under poor lighting conditions, as the model relies heavily on visual information to make predictions. Conversely, however, we observed that the accuracy of the state-of-the-art model was higher at night when lighting conditions were significantly poorer compared to sunny days. We found that the reason was because the state-of-the-art model was not able to make a decision at all due to extremely poor visibility of the pedestrian, and this ``no prediction'' by the model was counted as a prediction of ``not crossing''. As a result, most of the non-crossing actions were considered as being predicted correctly even though the model did not even make a prediction. Due to this counter-intuitive observation, we focused on the recall results since the recall value represents the percentage of correctly predicted events out of all pedestrian-crossing events. In Table~\ref{table:lighting}, the recall results indicate that the performance for the state-of-the-art model actually degraded under poor lighting conditions.

We made a similar observation for our model. The reason was because our model also exploited visual information to make a prediction although the decision was reinforced by incorporating motion sensor data. However, our results show that our model is significantly more robust to poor lighting conditions compared to the state-of-the-art model. More specifically, the recall results for our model indicate that our model performs consistently well in all lighting conditions. In particular, the performance degradation in terms of the recall value at night compared with that under the sunny condition was only 0.7\%, compared to 19.8\% for the state-of-the-art model. Additionally, we observed that our model achieved higher accuracy, AUC, F1-score, and precision compared with the state-of-the-art model in most lighting conditions. Our results demonstrate the significant advantages of integrating motion sensor data to improve model performance, highlighting the limited decision-making capacity of both vision and non-vision branches when used in isolation. By combining visual information with motion sensor data, our model was able to more effectively interpret pedestrian behavior and make accurate predictions, even under challenging scenarios such as poor lighting or long distances.



\section{Conclusions}
\label{sec:conclusion}

In this study, we introduced WatchPed, a novel framework for pedestrian's crossing intention prediction that addresses the limitations of state-of-the-art approaches, which often rely heavily on visual information to make predictions. By integrating motion sensor data and other features, our model was able to make more accurate and reliable predictions, even in challenging scenarios where visual information alone may be insufficient.  Additionally, we have performed a large-scale data collection process and introduced the first of its kind pedestrian's crossing intention dataset integrated with time-synchronized motion sensor data. The dataset consists of a total of 255 video clips and is strategically organized according to different distances and lighting conditions aiming to spark research on pedestrian intention prediction solutions based on the motion sensor data. We have trained and tested our model using the widely adopted JAAD dataset and our own dataset to demonstrate the effectiveness of our model compared with a state-of-the-art approach.
\bibliographystyle{IEEEtran}
\bibliography{mybibfile}

\end{document}